\documentclass{article}

\usepackage{arxiv}

\usepackage[utf8]{inputenc} % allow utf-8 input
\usepackage[T1]{fontenc}    % use 8-bit T1 fonts
\usepackage{hyperref}       % hyperlinks
\usepackage{url}            % simple URL typesetting
\usepackage{booktabs}       % professional-quality tables
\usepackage{amsfonts}       % blackboard math symbols
\usepackage{nicefrac}       % compact symbols for 1/2, etc.
\usepackage{microtype}      % microtypography
\usepackage{lipsum}
\usepackage{graphicx}

\usepackage{caption}
\usepackage{multirow}
\usepackage{adjustbox}
\usepackage{amsmath}
\usepackage{graphicx,subcaption} 
\usepackage{tabularx}
\usepackage{amssymb}
\usepackage{amsmath}
\usepackage{xcolor}
\usepackage[square,sort,comma]{natbib}
\usepackage{authblk}

\setcitestyle{numbers}

\title{LMS-AutoTSF: Learnable Multi-Scale Decomposition and Integrated Autocorrelation for Time Series Forecasting}

\author[1,2]{Ibrahim Delibasoglu}
\author[1]{Sanjay Chakraborty}
\author[1]{Fredrik Heintz}

\affil[1]{Department of Computer and Information Science (IDA), Linköping University, Linköping, Sweden}
\affil[2]{Software Engineering, Sakarya University, Sakarya, Turkiye}

\begin{document}
\maketitle
\begin{abstract}
Time series forecasting is an important challenge with significant applications in areas such as weather prediction, stock market analysis, and scientific simulations. In this work, we introduce LMS-AutoTSF, a novel time series forecasting architecture that incorporates autocorrelation while leveraging dual encoders operating at multiple scales. Unlike traditional models that rely on predefined trend and seasonal components, LMS-AutoTSF employs two separate encoders per scale: one focusing on low-pass filtering to capture trends and the other utilizing high-pass filtering to model seasonal variations. These filters are learnable, allowing the model to dynamically adapt and isolate trend and seasonal components directly in the frequency domain. A key innovation in our approach is the integration of autocorrelation, achieved by computing lagged differences in time steps, which enables the model to capture dependencies across time more effectively. Each encoder processes the input through fully connected layers to handle temporal and channel interactions. By combining frequency-domain filtering, autocorrelation-based temporal modeling, and channel-wise transformations, LMS-AutoTSF not only accurately captures long-term dependencies and fine-grained patterns but also operates more efficiently compared to other state-of-the-art methods. Its lightweight design ensures faster processing while maintaining high precision in forecasting across diverse time horizons. The source code is publicly available at \url{http://github.com/mribrahim/LMS-TSF}
\end{abstract}

% keywords can be removed
%\keywords{First keyword \and Second keyword \and More}

\twocolumn

\section{Introduction}
\label{sec-intro}
Time series data typically comprises an ordered sequence of observed or measured outcomes from a process at fixed time intervals. These datasets are designed to capture relevant information and activities within a specific subject matter. The primary goal of time series applications is to estimate future values by the identification of underlying patterns in historical data. Time series forecasting (TSF) is a frequently used technique in fields like traffic congestion anticipation, weather forecasting, and stock market prediction. Making more informed and efficient decisions is made possible by accurate forecasting, which gives decision-makers the ability to recognise and reduce risks \cite{li2024deep}. Due to the complex and ever-changing nature of real-world systems, observed time series often exhibit intricate temporal patterns. These patterns can involve a mix of increasing, decreasing, and fluctuating trends, making forecasting highly challenging. Multivariate time series forecasting involves predicting future values of multiple interrelated time-dependent variables based on their historical data. Unlike univariate forecasting, which deals with a single variable, multivariate forecasting considers the interactions and dependencies between multiple variables, allowing for more complex and potentially more accurate predictions across several dimensions \cite{zerveas2021transformer}. In time series data, the trend component represents the long-term directional movement, showing consistent upward or downward patterns, while the seasonal component captures regular, repeating fluctuations over fixed intervals, such as daily, monthly, or yearly cycles. Variations in trends can be linear or nonlinear and may shift or reverse over time, whereas seasonal variations are typically periodic but can vary in amplitude or timing due to external factors. Both components often overlap, making the task of isolating and forecasting them more complex \cite{chen2023long}. In time series analysis, frequency domain filtering has two types. A low-pass filtering is commonly used to capture seasonal variations by allowing low-frequency components of the data to pass through while attenuating high-frequency noise. This process helps isolate recurring seasonal patterns, such as annual or monthly cycles, which are important in fields like finance and climatology \cite{hyndman2018forecasting}. Conversely, high-pass filtering is employed to model trends by passing high-frequency components and filtering out lower-frequency signals, which represent long-term, smooth variations. This technique is particularly useful for identifying short-term fluctuations and trends, removing the impact of long-term cyclical patterns \cite{shumway2000time}. Both methods provide a clearer decomposition of time series data, enabling analysts to focus on specific components of interest. Transformer is emerging in time series forecasting, driven by the tremendous success in the natural language processing field. Transformer has the ability to extract multi-level representations from sequences and illustrate pairwise connections \cite{liu2023itransformer}. In time series data analysis, Transformers that combine frequency-domain filtering with temporal and channel-wise transformations offer a powerful approach to capturing complex dependencies and patterns. By leveraging frequency-domain filtering, these models can efficiently separate and analyze the signal's various frequency components, isolating key seasonal or trend-related elements \cite{zerveas2021transformer}. Additionally, temporal and channel-wise transformations allow the model to process both the sequential nature of time series data and the relationships across multiple channels, making it particularly adept at handling multivariate time series and discovering intricate dependencies between variables \cite{zhang2022less}. This dual capability enables the model to address both local and global patterns, improving performance in tasks such as forecasting, anomaly detection, and classification. Our main contributions are as follows: 
\begin{itemize}
    \item The dynamic decomposition enhances the model's ability to capture intricate patterns and the model can learn the trend and seasonality features of each dataset without depending on fixed assumptions.
    \item  By leveraging the concept of autocorrelation in the time domain, the architecture effectively incorporates temporal dependencies into the forecasting process. This integration allows the model to recognize and utilize historical relationships within the data, improving forecasting accuracy.
    \item LMS-AutoTSF achieves state-of-the-art performance in time series forecasting tasks by integrating learnable decomposition, autocorrelation, and multi-scale processing across a wide range of benchmarks.
    \item LMS-AutoTSF is a lightweight and computationally efficient architecture, that significantly reduces processing time compared to existing state-of-the-art methods, while maintaining or even improving forecasting accuracy.
\end{itemize}
We do comprehensive tests on different time-series forecasting benchmarks to support our motivation and hypothesis. When compared to various forecasting techniques, the performance of our proposed model is at the cutting edge. Extensive ablation investigations and analysis trials support the viability of suggested designs and their consequent benefit over earlier methods.\\
This paper is organized as follows. Section \ref{sec-literature} explains a set of typical times series forecasting works from the literature which sets the notion and motivation of this paper. Section \ref{sec-problem} briefly explains the problem statement of this paper. In Section \ref{sec-methodology}, we provide a detailed discussion of our proposed architecture called LMS-AutoTSF which is suitable for multivariate time series forecasting. In section \ref{sec-experiments}, we discuss the result analysis of the proposed methodology and describe a detailed comparison with the state-of-the-art time series forecasting models. A conclusion of the work's findings is provided in Section \ref{sec-conclusion}.

\section{Literature Review}
\label{sec-literature}
In long-term and short-term time series forecasting, transformers have demonstrated exceptional performance and significant potential \cite{zerveas2021transformer,zhang2024multivariate,nie2022time,zeng2023transformers}. Informer \cite{zhou2021informer}, the first renowned transformer for time-series forecasting, uses a generative style decoder and ProbSparse self-attention to solve problems like quadratic time complexity. A number of models were introduced after Informer\cite{zhou2021informer}, including Autoformer \cite{chen2021autoformer}, Pyraformer \cite{liu2021pyraformer}, iTransformer \cite{liu2023itransformer}, Reformer \cite{kitaev2020reformer} and FEDFormer \cite{zhou2022fedformer}. While Pyraformer\cite{liu2021pyraformer} concentrates on multiresolution attention for signal processing efficiency. Autoformer\cite{chen2021autoformer} employs auto-correlation and decomposition for performance, while FEDFormer \cite{zhou2022fedformer} combines frequency analysis with Transformers for improved time series representation. By emphasising the value of patches, the PatchTST \cite{nie2022time} improves the model's capacity to identify both local and global relationships in data. To leverage the cross-dimension dependency, a Crossformer \cite{zhang2023crossformer} model is presented for multivariate time-series forecasting. The Dimension-Segment-Wise (DSW) embedding approach is used in Crossformer to embed the input time-series data into a 2D vector array while preserving time and dimension information. Thus, the Two-Stage Attention (TSA) layer successfully captures the cross-time and cross-dimension dependency. Using the DSW embedding and TSA layer, Crossformer builds a Hierarchical Encoder-Decoder (HED) that utilises the data at different scales for the ultimate prediction. A multilayer perceptron (MLP) architecture is used by the time series forecasting model TSMixer \cite{ekambaram2023tsmixer} to blend and process temporal information. Token-mixing layers, which are used by TSMixer in place of conventional recurrent or convolutional neural networks, capture dependencies between several time steps and feature dimensions. TSMixer is able to forecast across a variety of time series datasets with efficiency and effectiveness by concentrating on these relationships. Its straightforward yet effective architecture enables parallel processing, which reduces training and inference times while preserving competitive performance when compared to more intricate models like Transformers and LSTMs. TimeMixer \cite{wang2024timemixer}, a complete multi-layer perceptron-based architecture, takes the entire benefit of disentangled multiscale time-series in both the past extraction and future prediction stages. The Past-Decomposable-Mixing (PDM) and Future-Multipredictor-Mixing (FMM) components make up this system. Specifically, PDM uses the decomposition to multiscale series and then independently combines the seasonal and trend patterns that have been decomposed in fine-to-coarse and coarse-to-fine directions, therefore aggregating the macroscopic trend information and the microscopic seasonal information in turn. FMM additionally groups several predictors to take use of their complimentary predicting skills in multiscale observations. For time series forecasting, the DLinear model \cite{zeng2023transformers} is a simple and effective method that breaks down a time series into trend and seasonal components. By modelling each component independently using linear layers, DLinear streamlines the forecasting process in contrast to intricate deep learning models. The model is able to extract from the data both short-term seasonal patterns and long-term trends thanks to this decomposition. Besides transformer models, a novel norm-bounded graph attention network (GAT) is generated for multivariate time-series forecasting by upper-bounding the Frobenius norm of weights in each layer of the GAT model to improve performance and address the basic over-smoothing issue in deep graph neural networks-based models \cite{zhang2024multivariate}. The problem is made more complex by GAT's over-smoothing, which is more complex because it multiplies several attention matrices at different moments. GAT's over-smoothing is less obvious than other methods. The trainable adaptive parameters in weight-bound control may be able to reduce over-smoothing by distributing the attention matrices throughout layers and nodes. Autoformer and FEDFormer are transformer-based methods that rely on attention mechanisms to capture long-term dependencies and decompose time series into trend and seasonal components. Autoformer employs progressive attention and learnable decomposition, while FEDFormer integrates frequency domain features and federated attention for improved efficiency. TimeMixer, in contrast, adopts a non-transformer-based approach similar to ours, utilizing mixing components for representation learning. These methods share some conceptual similarities with our architecture, such as leveraging decomposition techniques and multi-scale processing. However, our lightweight, fully connected architecture relies on autocorrelation as a core feature in skip connection and FFT-based learnable filters to extract trend and seasonal components effectively. This streamlined design achieves superior performance while remaining computationally efficient, as evidenced by our experimental results.

\begin{figure*}[htb]
\centering
\includegraphics[scale=0.2]{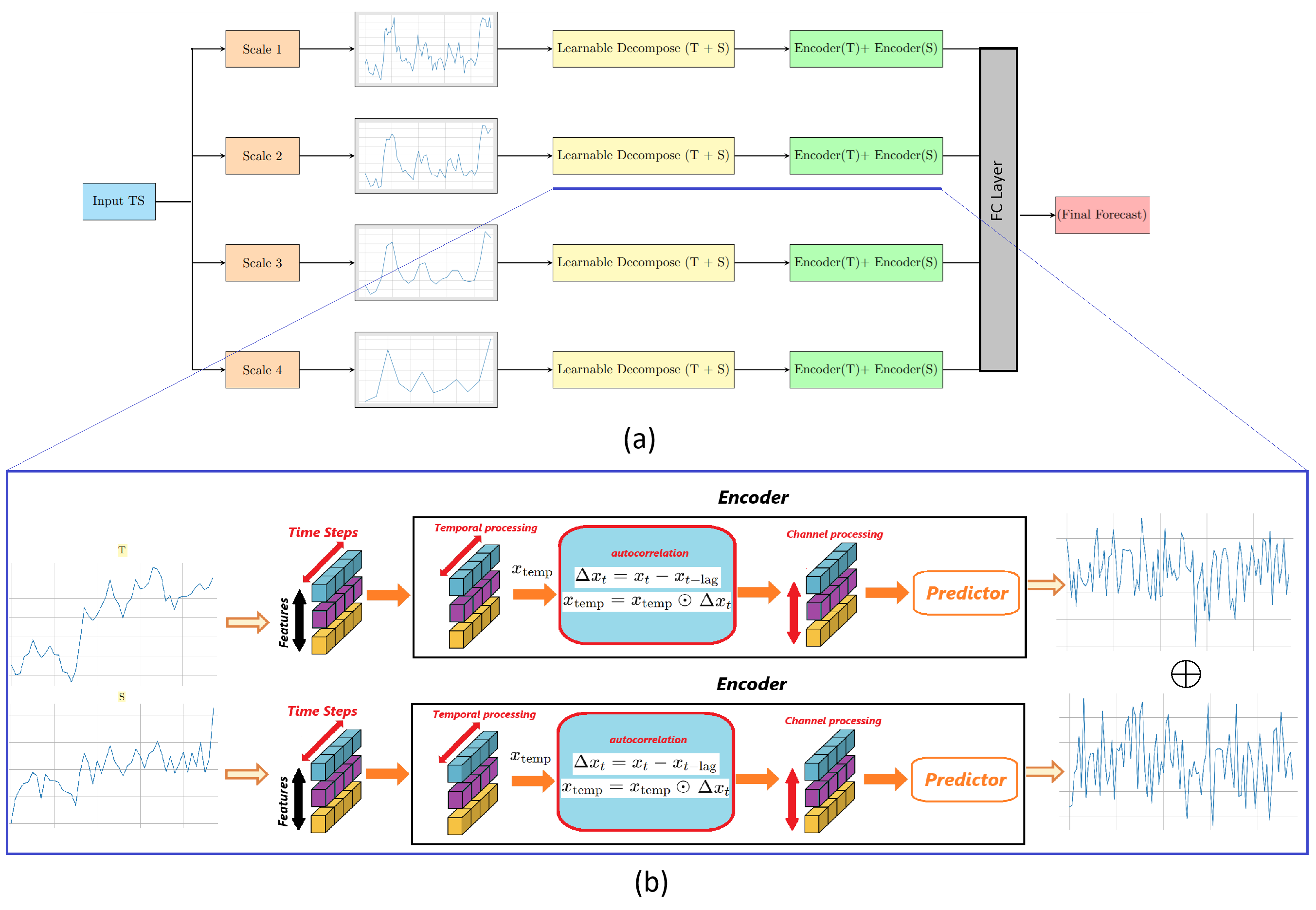}
\caption{ (a) Overall architecture of LMS-AutoTS. (b) Encoder module operations for trend and seasonal components in Scale-2}
\label{fig-overview}
\end{figure*}

\section{Problem Statement}
\label{sec-problem}

This paper addresses the challenge of long-term forecasting for multivariate time series, using historical data. We define a multivariate time series at time \( t \) as \( \mathbf{X}_t = [x_{t,1}, x_{t,2}, \dots, x_{t,N}] \), where \( x_{t,n} \) represents the value of the \( n \)-th variable at time \( t \), for \( n = 1, 2, \dots, N \). We use the notation $X_{t:t+h}$ to represent the $X$-values from the time point $t$ to $t+H$, inclusive. The aim is to develop a model for forecasting the future values of the series over the next \( T \) time steps, based on the most recent \( L \) time steps. The parameters \( L \) and \( H \) are referred to as the \textit{look-back window} and the \textit{prediction horizon}, respectively.
Specifically, for a given initial time \( t_0 \), the model takes as input the sequence \( \mathbf{X}_{t_0-L:t_0} \), corresponding to the past \( L \) time steps, and outputs the predicted sequence \( \hat{\mathbf{X}}_{t_0:t_0+H} \), representing the forecasted values for the next \( H \) time steps. The predicted value of \( \mathbf{X}_t \) at time \( t \) is denoted by \( \hat{\mathbf{X}}_t \). In brief, the goal of multivariate time series forecasting is to predict future values \( \mathbf{X}_{t:t+H} \) given past observations \( \mathbf{X}_{t-L:t} \):
\begin{equation}
\hat{\mathbf{X}}_{t:t+H} = f(\mathbf{X}_{t-L:t})
\end{equation}
The forecasting performance of the model is assessed by computing the mean squared error (MSE) and the mean absolute error (MAE) between the prediction and the ground truth on the test set. For the M4 datasets, we use the evaluation methodology introduced by N-BEATS and use mean absolute percentage error (MAPE), symmetric MAPE (sMAPE), mean absolute scaled error (MASE), and overall weighted average (OWA) as evaluation metrics.

\section{Methodology}
\label{sec-methodology}

In this work, we introduce a novel approach for time series forecasting that leverages a combination of multi-scale input processing, frequency domain filtering \cite{yi2024frequency}, and autocorrelation. The general overview of the proposed architecture is represented in Figure \ref{fig-overview}. To capture both local and global patterns in the time series, we apply multi-scale down-sampling of the input data. The input sequence is progressively down-sampled using an average pooling layer, with each scale indexed by $k$ representing different temporal resolutions. This enables the model to capture variations across multiple time horizons. In the rest of the article, we use the abbreviation $X_t^{LB}$ to represent $X_{t-L:t}$, where "LB" stands for "loopback".

\begin{equation}
\mathbf{X}_{t:t+H}^{(k)} = \text{Downsample}_k(\mathbf{X}_t^{LB})
\end{equation}

For each scale \( k \), downsampled time series is decomposed into trend (\( T \)) and seasonality (\( S \)) components with learnable (differentiable) layers in the frequency domain.  The decomposition is achieved by applying a frequency-domain transformation (via FFT), followed by learnable low-pass and high-pass filters. The cutoff frequency and steepness of the filters are trainable parameters, and different filters are learned for each feature of the time series. The filtering process is defined as:
\begin{itemize}
    \item \textbf{Frequency transform}:  Fast Fourier Transform (FFT) efficiently transforms a time series $X_t$ from the time domain into its frequency domain representation $X(f)$, and vice versa, by decomposing the series into a sum of sinusoidal components.
    
\begin{equation}
\text{FFT}(X_{\text{LB}}) = \sum_{i=0}^{L-1} X_{t-L+i} \cdot e^{-j \frac{2 \pi i}{L}}
\end{equation}

\begin{equation}
     X_{freq} = FFT(X_t^{LB})
\end{equation}
    \item \textbf{Low-pass filter}: Captures the trend \textbf{(T)} by retaining lower frequencies, defined as:
\begin{equation}
    T = X_{\text{low}} = \text{FFT}^{-1}(X_{\text{freq}} \cdot \sigma(-(f - f_{\text{cutoff}}) \cdot s))
\end{equation}
    where $f$ is the frequency, $f_{\text{cutoff}}$ is the learnable cutoff frequency, $s$ is the steepness, and $\sigma$ is the sigmoid function.
    
    \item \textbf{High-pass filter}: Extracts the seasonal component \textbf{(S)} by retaining higher frequencies, defined similarly:
\begin{equation}
    S = X_{\text{high}} = \text{FFT}^{-1}(X_{\text{freq}} \cdot \sigma((f - f_{\text{cutoff}}) \cdot s))
\end{equation}
\end{itemize}
The low-pass filter is applied to emphasize low-frequency components (i.e., trend), while the high-pass filter is used to isolate high-frequency components (i.e., seasonality). This approach allows for the effective decomposition of the time series into its underlying trend and seasonal components, improving forecasting accuracy. This trainable decomposition approach enables the model to learn trend and seasonality components directly from the data, rather than relying on a predefined method. This flexibility allows the model to capture patterns that might be missed by traditional techniques, resulting in more precise forecasts. Furthermore, this adaptability supports better generalization across different datasets, as the model can learn the unique trend and seasonality features of each dataset without depending on fixed assumptions.
\begin{equation}
\mathbf{X}_t^{LB}{(k)} = T^{(k)} + S^{(k)}
\end{equation}
Encoder module details are represented in Figure \ref{fig-overview} (b). In the Encoder module, we apply fully connected ($FC$) layers for temporal and channel processing. In addition, we apply autocorrelation via lagged difference, so the model takes this feature into account in addition to trend or season to emphasize how much temporal variation matter for forecasting. Autocorrelation in the time domain refers to the correlation between a time series and its lagged version, capturing how past values influence future values. A simple way to compute this is by taking the difference between a time series at time $t$ and its value at a previous time $t-1$, which highlights how much the time series changes over consecutive steps. By incorporating autocorrelation in the model, the network can more effectively learn temporal dependencies, improving its ability to forecast future values based on past behavior. 
In the Encoder, firstly input (consider $T$ for scale $k$: ${T}_t^{(k)}$) is passed through a temporal processing $FC$ layer, and then processed temporal data $x_{temp}$ is multiplied by the autocorrelation which is the lagged difference of the input of the Encoder module as shown below.
\begin{equation}
    x_{temp} = \text{FC}_{\text{temp}}(T^{(k)})
\end{equation}
\begin{equation}
\begin{split}
x_{temp} &= x_{temp} \odot \Delta T^{(k)}
\end{split}
\end{equation}
where \( \odot \) represents element-wise multiplication. Then, channel processing operation is performed to handle feature interactions, and it is followed by a final layer to produce the final forecasting output: $\hat{T}_{t}^{(k)}$ for forecasting from $T$, $\hat{S}_{t}^{(k)}$ for forecasting from $S$. 
\begin{equation}
\begin{split}
    x_{channel} = \text{FC}_{\text{channel}}(x_{temp}) \\
    \hat{T}^{(k)} = \text{FC}_{\text{projection}}(x_{temp} + x_{channel})
\end{split}
\end{equation}
The encoder module is applied to forecast from trend and seasonal components separately, and each forecast is summed to find the final forecast output of the corresponding scale as shown with \( \oplus \) operation in Figure \ref{fig-overview}. The forecasting outputs of encoder for trend and seasonality components are summed to form the final output for each scale:
\begin{equation}
\begin{split}
\hat{T}^{(k)} = \text{Encoder}_T^{(k)}(T^{(k)}) \\
\hat{S}^{(k)} = \text{Encoder}_S^{(k)}(S^{(k)})
\end{split}
\end{equation}
\begin{equation}
\mathbf{\hat{X}}_{t:t+H}^{(k)} = \hat{T}^{(k)} + \hat{S}^{(k)}
\end{equation}
The final prediction \( \mathbf{\hat{X}}_{t:t+H} \) is obtained by applying a final $FC$ layer to the concatenated forecasts from all scales \( k \):
\begin{equation}
\mathbf{\hat{X}}_{t:t+H} = \text{FC} \left( \left[ \mathbf{\hat{X}}_{t:t+H}^{(1)} \; \mathbf{\hat{X}}_{t:t+H}^{(2)} \; \cdots \; \mathbf{\hat{X}}_{t:t+H}^{(K)} \right] \right)
\end{equation}
Where:
\( \left[ \mathbf{\hat{X}}_{t:t+H}^{(1)} \; \mathbf{\hat{X}}_{t:t+H}^{(2)} \; \cdots \; \mathbf{\hat{X}}_{t:t+H}^{(K)} \right] \) represents the concatenation of forecasts from all scales, - \( K \) is the total number of scales.

\section{Experiments}
\label{sec-experiments}
In this section, we perform in-depth experiments to assess how well our proposed LMS-AutoTSF model forecasts in conjunction with state-of-the-art time-series forecasting architectures. An ablation work is also applied to measure the effect of the proposed modules of learnable decomposition and autocorrelation. All the experiments are implemented in PyTorch, CUDA version 12.2 and conducted on a single NVIDIA-GeForce RTX 3090 with 24GB GPU. We have replicated all of the compared baseline models and implemented them using the benchmark Time-Series Library (TSLib) \cite{wu2022timesnet} repository, which is based on the configurations provided by the official code or actual article for each model. We train the proposed model on all datasets using a batch size of $32$, a learning rate of $0.0001$ except PEMS, and the ADAM optimizer with L2 loss. The model operates with $K=4$ scales.

\begin{table*}[]
\centering
\scriptsize
\caption{Ablation study for the average results of prediction horizons: {96, 192, 336,720}}
\begin{tabular}{|cc|c|ll|ll|ll|ll|}
\hline
\textbf{\begin{tabular}[c]{@{}c@{}}Fixed \\ decompose\end{tabular}} & \textbf{\begin{tabular}[c]{@{}c@{}}Learnable \\ decompose\end{tabular}} & \textbf{\begin{tabular}[c]{@{}c@{}}Auto\\ correlation\end{tabular}} & \multicolumn{2}{c|}{\textbf{Etth1}} & \multicolumn{2}{c|}{\textbf{Ettm1}} & \multicolumn{2}{c|}{\textbf{Weather}} & \multicolumn{2}{c|}{\textbf{Electricity}} \\ \hline
\multicolumn{1}{|l}{}                                               & \multicolumn{1}{l|}{}                                                   & \multicolumn{1}{l|}{}                                               & \textbf{MSE}     & \textbf{MAE}     & \textbf{MSE}     & \textbf{MAE}     & \textbf{MSE}      & \textbf{MAE}      & \textbf{MSE}        & \textbf{MAE}        \\
\checkmark                                           & $-$                                                                     & $-$                                                                 & 0.457            & 0.440            & 0.398            & 0.406            & 0.242             & 0.273             & 0.180               & 0.279               \\
$-$                                                                 & \checkmark                                               & $-$                                                                 & 0.448            & 0.435            & 0.392            & 0.402            & 0.240             & 0.270             & 0.174               & 0.271               \\
$-$                                                                 & \checkmark                                               & \checkmark                                           & 0.441            & 0.432            & 0.377            & 0.394            & 0.238             & 0.269             & 0.175               & 0.271               \\ \hline
\end{tabular}
\label{table-ablation}
\end{table*}

\subsection{Datasets}
We evaluate the performance of our proposed architecture on eight widely-used public benchmark datasets, including Weather, Electricity, Traffic, Exchange and four ETT datasets (ETTh1, ETTh2, ETTm1, ETTm2) for long-term forecasting as detailed in supplemmentary file. The number of features in these datasets varies significantly, with Electricity containing 321 features and Traffic comprising 862 features, making them notably larger and more complex compared to others. In contrast, Weather has 21 features, Exchange has 8 features, and the four ETT datasets each have 7 features, allowing for a diverse evaluation of our architecture across datasets with varying feature scales.
The PEMS dataset \cite{chen2001freeway}, which is used for short-term traffic flow forecasting, includes four public traffic network datasets: PEMS03, PEMS04, PEMS07, and PEMS08. It provides an additional challenge with its highly dynamic and rapidly changing time series. Additionally, the M4 dataset includes 100,000 time series spanning various domains such as finance, industry, and demographics, with frequencies ranging from hourly to yearly, providing a comprehensive benchmark for both short-term and long-term forecasting tasks. These dataset allows us to evaluate the proposed model's ability to handle short-term prediction tasks with high variability. Our experiments show that transformer-based methods such as iTransformer and PatchTST struggle in many cases with the PEMS dataset, while our architecture demonstrates strong performance in addressing this challenging problem. Furthermore, our method achieves competitive performance compared to TimeMixer, which performs slightly better for short-term forecasting. However, our architecture outperforms TimeMixer in long-term forecasting tasks, highlighting its effectiveness across varying time horizons.

\subsection{Forecasting Results}
\begin{figure*}[]
  \begin{subfigure}{0.89\columnwidth}
  \includegraphics[width=\textwidth]{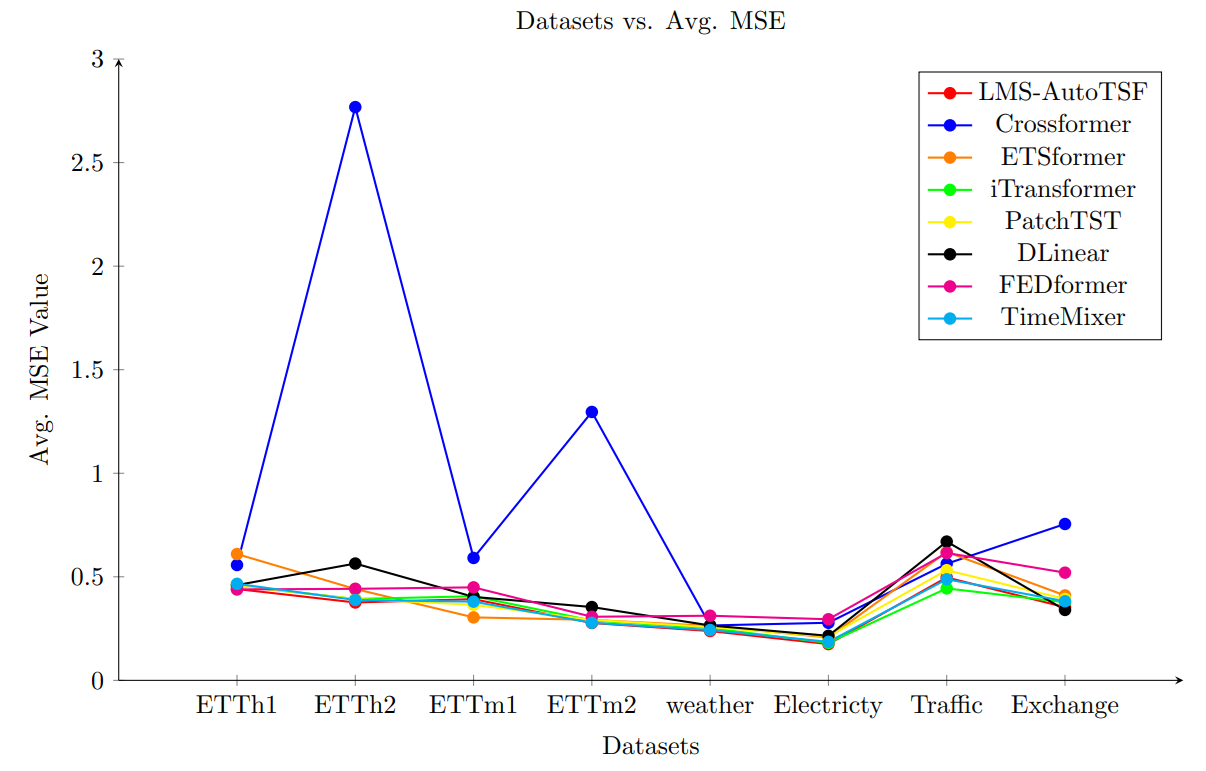}
  %\caption{Image A}
  \end{subfigure}
  \hfill
  \begin{subfigure}{0.89\columnwidth}
  \includegraphics[width=\textwidth]{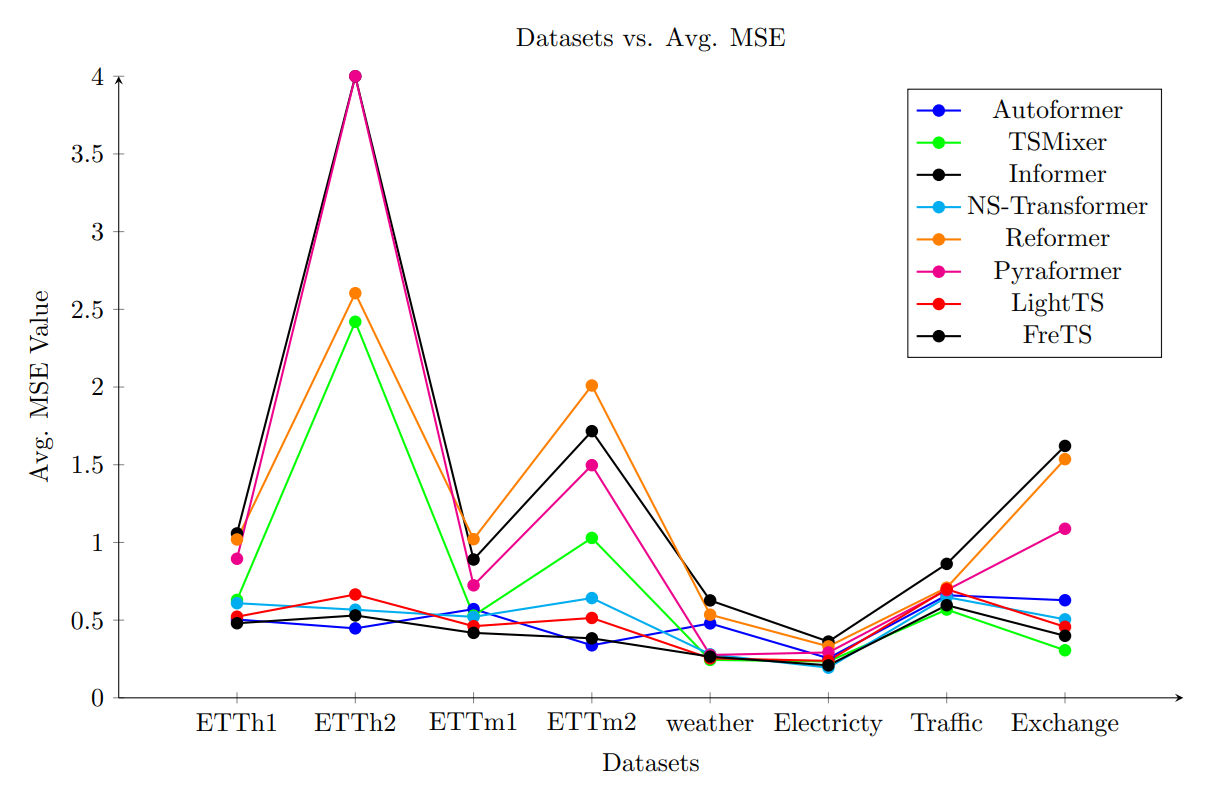}
  %\caption{Image B} 
  \end{subfigure} 
  \begin{subfigure}{0.89\columnwidth} 
  \includegraphics[width=\textwidth]{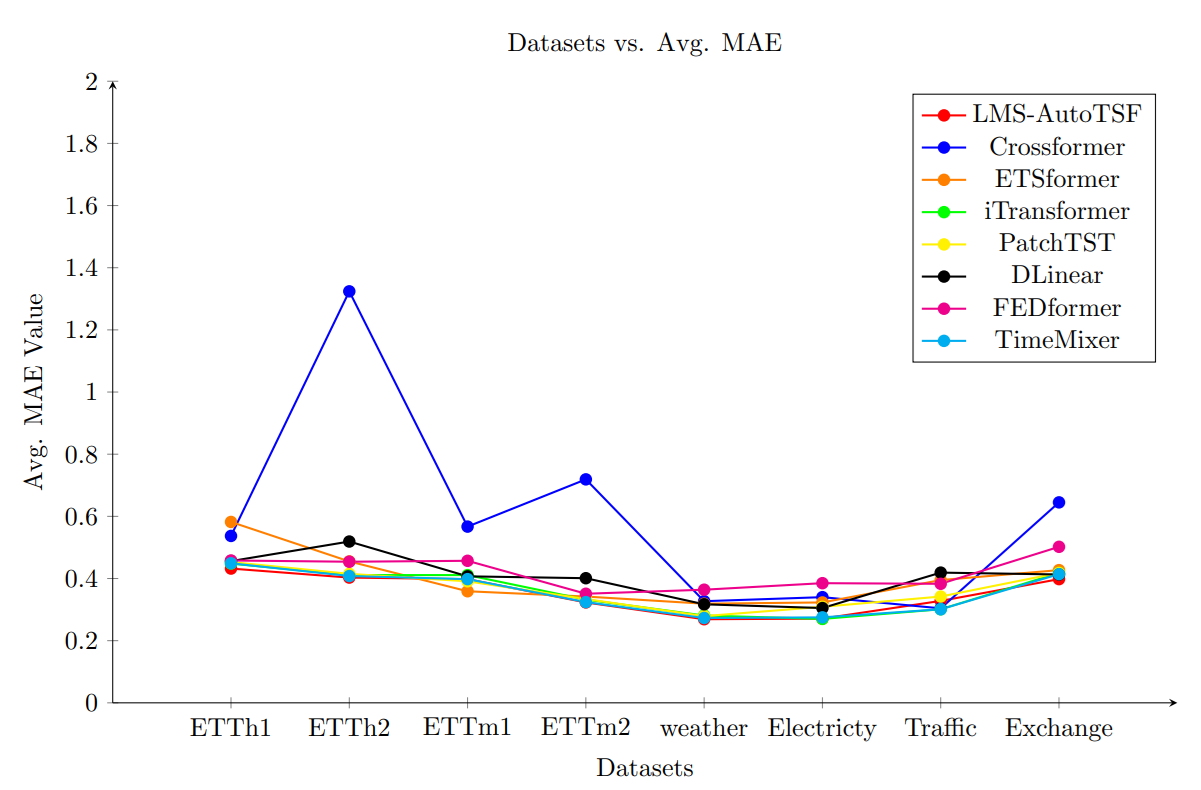} 
  %\caption{Image C} 
  \end{subfigure}  
  \hfill 
  \begin{subfigure}{0.89\columnwidth} 
  \includegraphics[width=\textwidth]{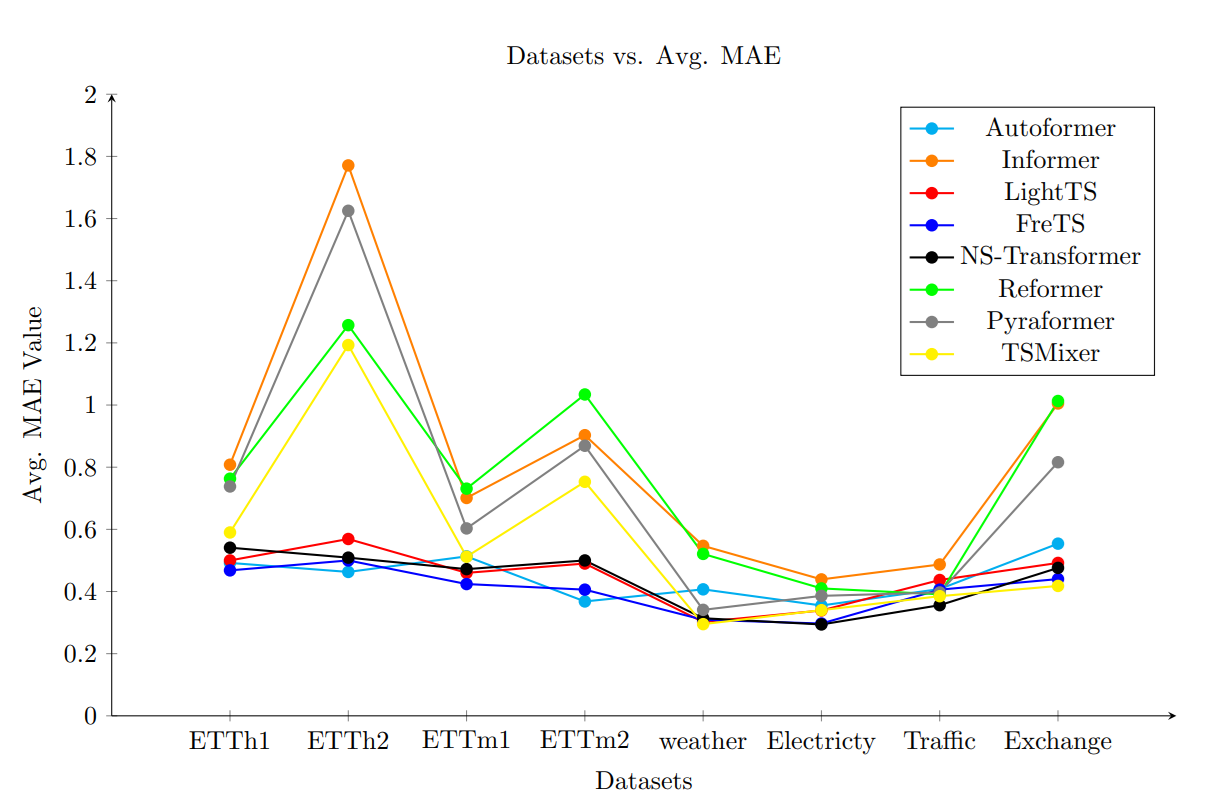} 
  %\caption{Image A again} 
  \end{subfigure}
  \caption{Comparison of models efficiency with datasets vs. avg. MSE vs. avg. MAE}
  \label{fig-comparison}
  \end{figure*}

\begin{figure}[]
  \begin{subfigure}{0.49\columnwidth}
  \includegraphics[width=\textwidth]{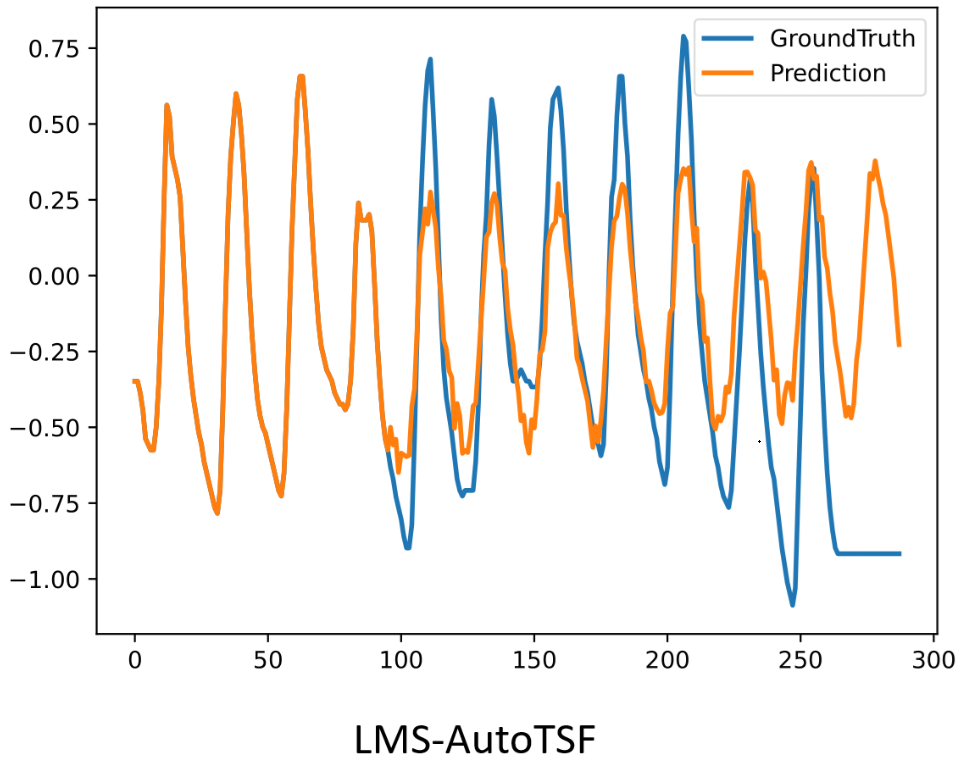}
  %\caption{Image A}
  \end{subfigure}
  \hfill
  \begin{subfigure}{0.49\columnwidth}
  \includegraphics[width=\textwidth]{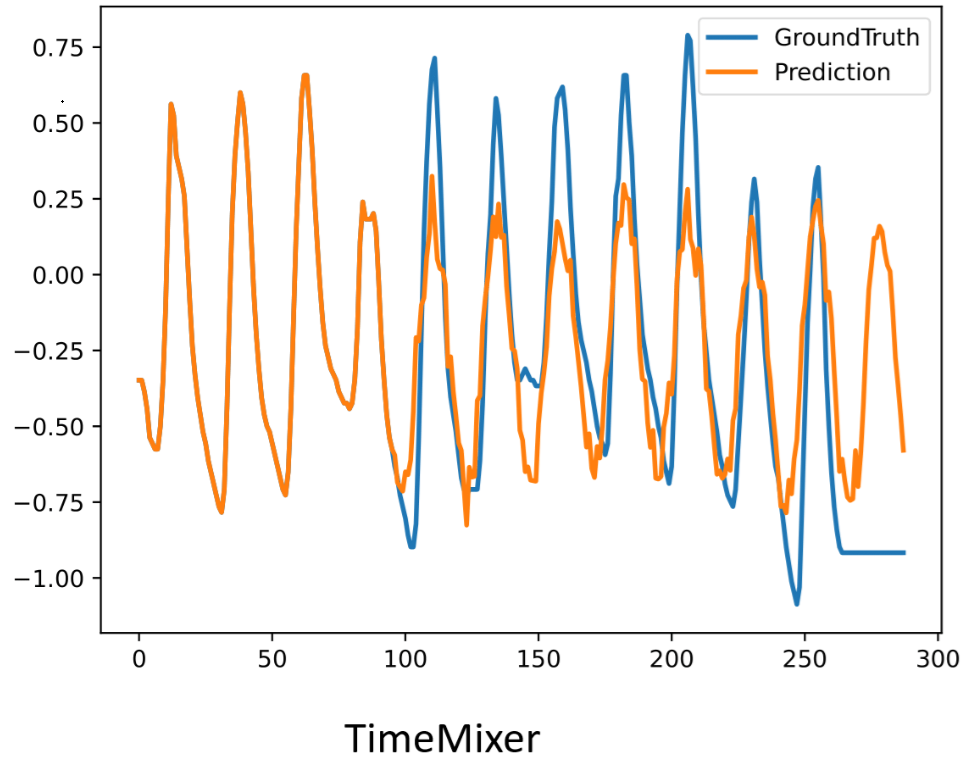}
  %\caption{Image B} 
  \end{subfigure} 
  \begin{subfigure}{0.49\columnwidth} 
  \includegraphics[width=\textwidth]{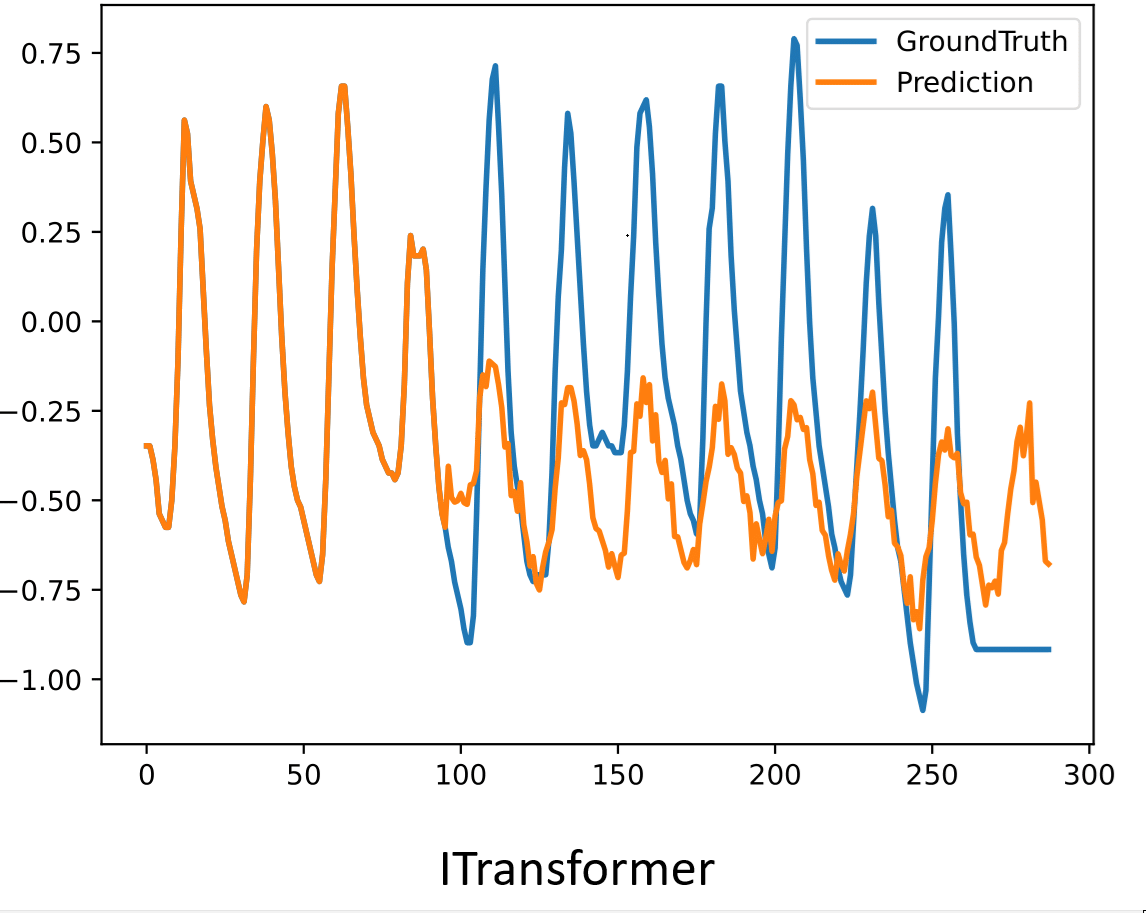} 
  %\caption{Image C} 
  \end{subfigure}  
  \hfill 
  \begin{subfigure}{0.49\columnwidth} 
  \includegraphics[width=\textwidth]{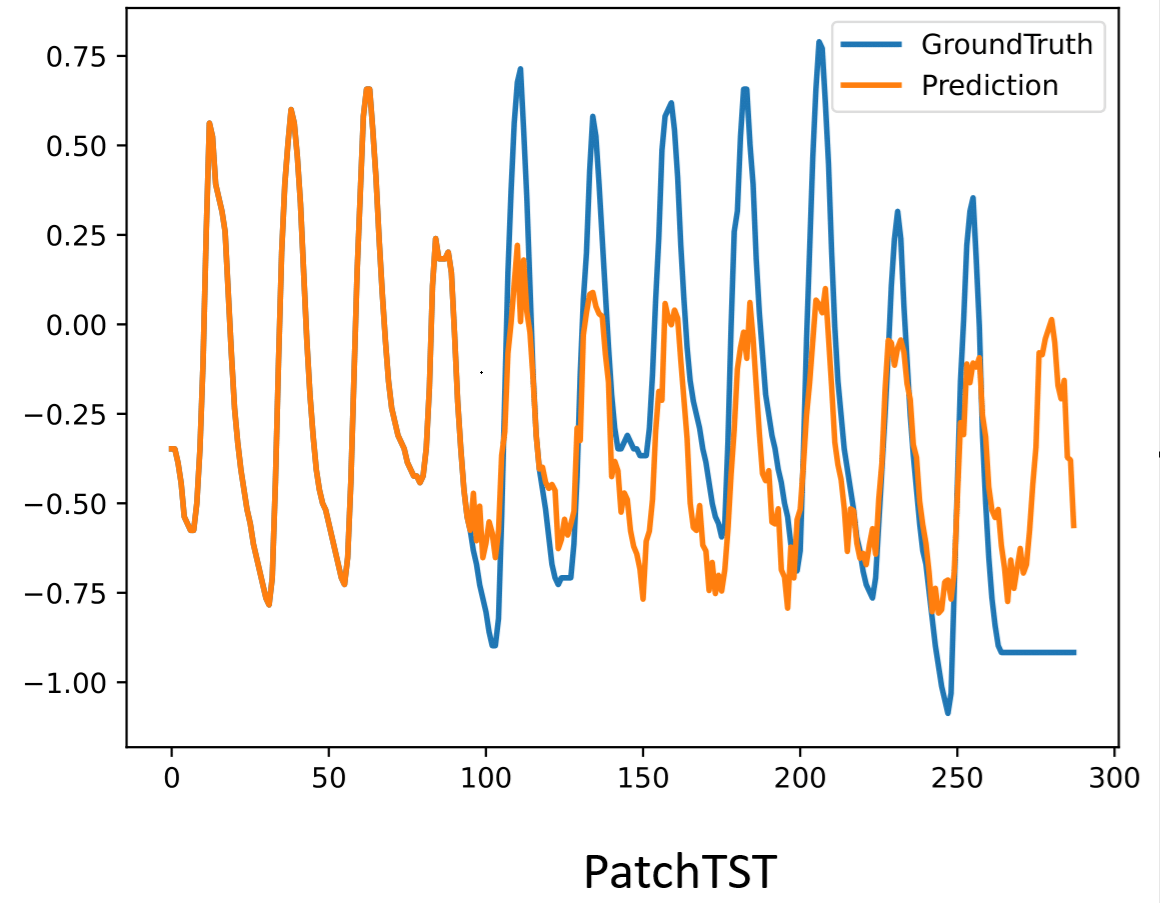} 
  %\caption{Image A again} 
  \end{subfigure}
  \caption{Visualization of prediction results on ETTh2 dataset}
  \label{fig-ETTh2-compare}
  \end{figure}

\begin{figure}[hbt!]
  \begin{subfigure}{0.49\columnwidth}
  \includegraphics[width=\textwidth]{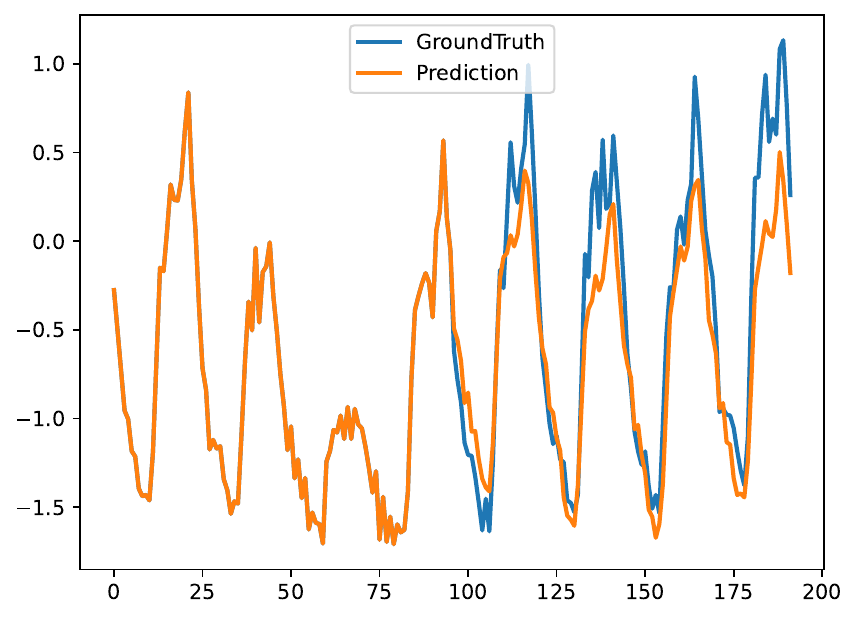}
   \caption{LMS-AutoTSF}
  \end{subfigure}
  \hfill
  \begin{subfigure}{0.49\columnwidth}
  \includegraphics[width=\textwidth]{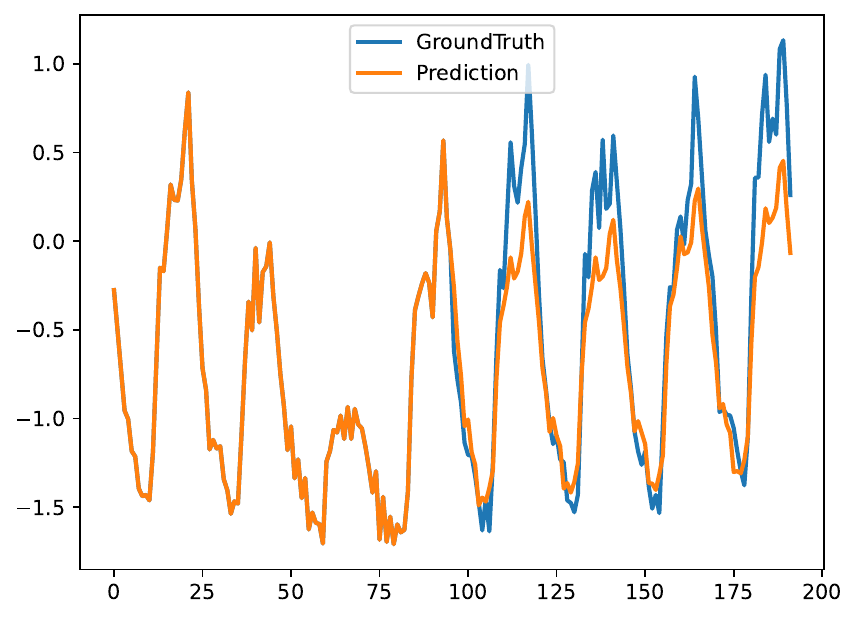}
  \caption{TimeMixer} 
  \end{subfigure} 
  \begin{subfigure}{0.49\columnwidth} 
  \includegraphics[width=\textwidth]{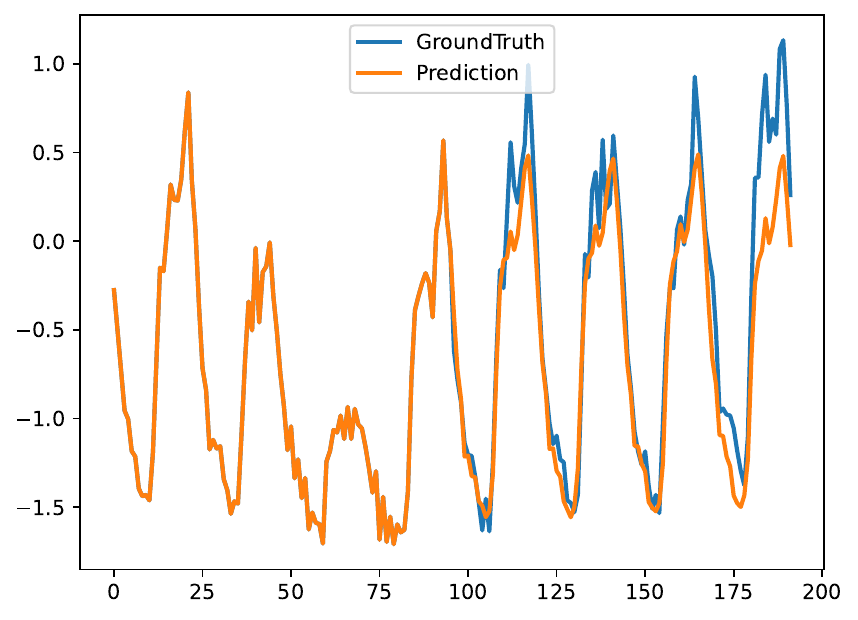} 
  \caption{iTransformer} 
  \end{subfigure}  
  \hfill 
  \begin{subfigure}{0.49\columnwidth} 
  \includegraphics[width=\textwidth]{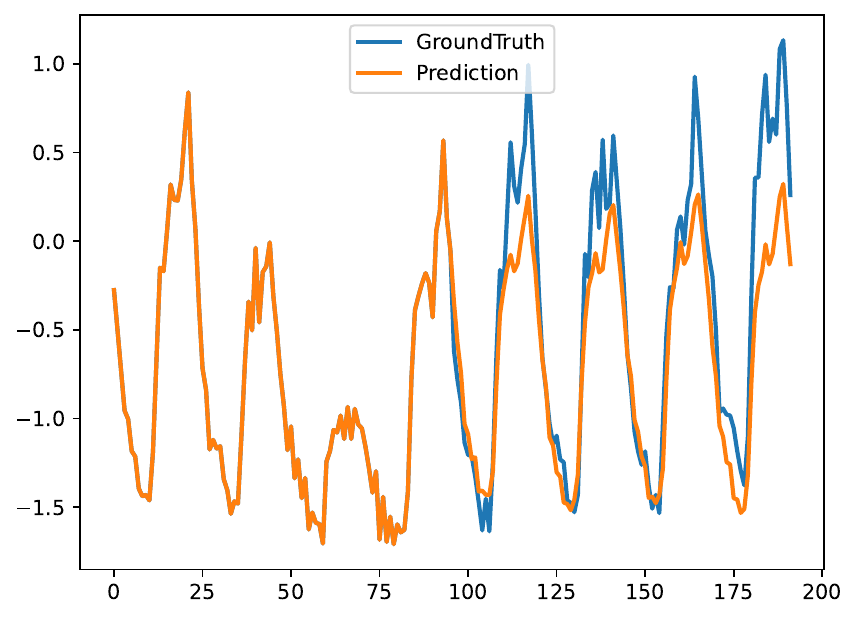} 
  \caption{PatchTST} 
  \end{subfigure}
  \caption{Visualization of prediction results on Electricity dataset}
  \label{fig-ECL-compare}
  \end{figure}

\begin{figure}[hbt!]
  \begin{subfigure}{0.49\columnwidth}
  \includegraphics[width=\textwidth]{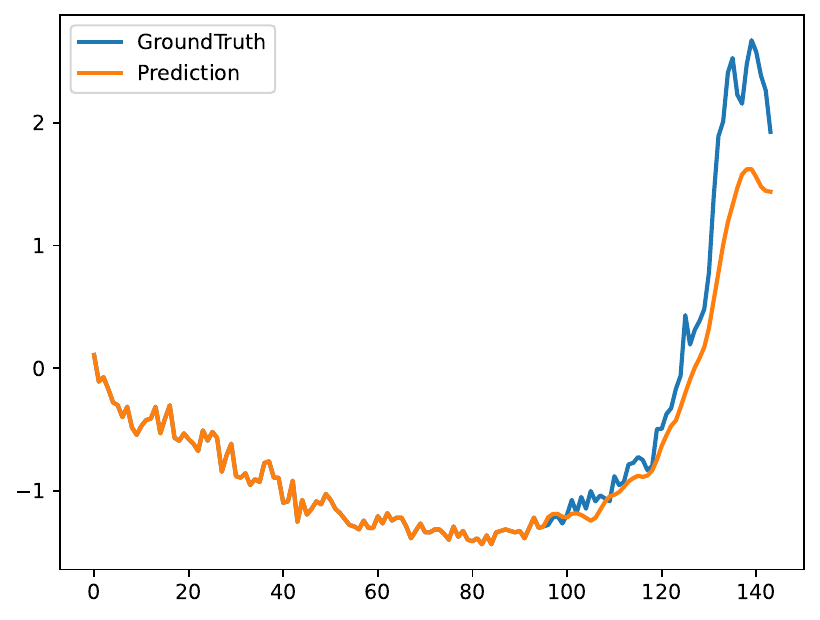}
  \caption{LMS-AutoTSF} 
  \end{subfigure} 
  \hfill
  \begin{subfigure}{0.49\columnwidth} 
  \includegraphics[width=\textwidth]{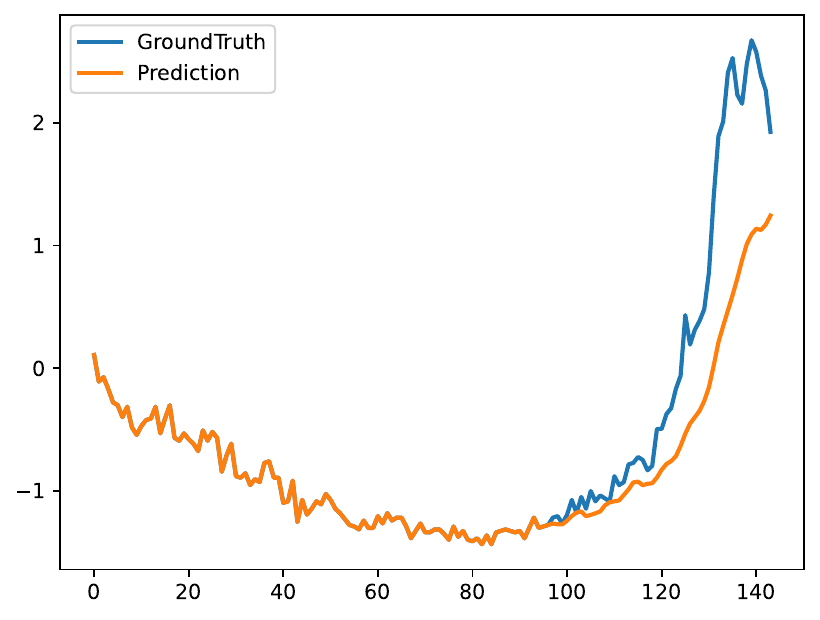} 
  \caption{TimeMixer} 
  \end{subfigure}
  \begin{subfigure}{0.49\columnwidth}
  \includegraphics[width=\textwidth]{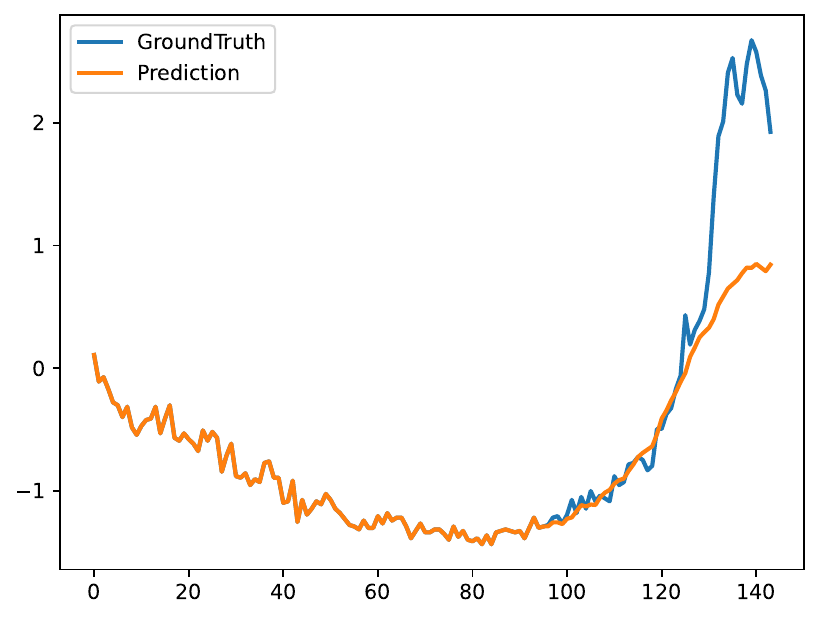}
  \caption{iTransformer}
  \end{subfigure}
  \hfill
  \begin{subfigure}{0.49\columnwidth} 
  \includegraphics[width=\textwidth]{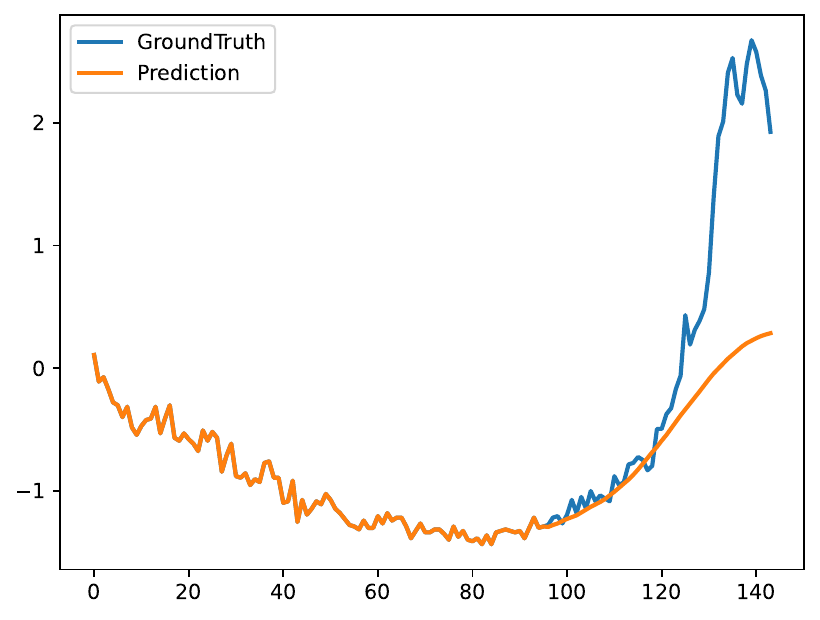} 
  \caption{PatchTST} 
  \end{subfigure}  
  \hfill 
  
  \caption{Visualization of prediction results on PEMS03 dataset}
  \label{fig-PEMS03-compare}
\end{figure}

Table \ref{table-ablation} provides insights into the contributions of various components of the LMS-AutoTSF architecture. Each row represents a different configuration of the model, showcasing the impact of using fixed decompomposition, learnable decomposition, and autocorrelation on forecasting performance.  The inclusion of autocorrelation further optimizes the model's performance, as seen in the third row, where both learnable decomposition and autocorrelation yield the lowest MSE and MAE values across most datasets. This suggests that integrating autocorrelation helps the model capture temporal dependencies more effectively. Comprehensive forecasting results are listed in Tables \ref{table-results}, \ref{table-resultsM4} and \ref{table-resultspems}. A lower MSE/MAE reflects more accurate predictions, and our proposed LMS-AutoTSF consistently achieves the best performance across most datasets for both long-term and short-term forecasting issue. It outperforms state-of-the-art models such as TimeMixer, iTransformer, and PatchTST, particularly excelling in high-dimensional time series forecasting. We have used sMAPE, MAPE, OWA, and MASE to evaluate the performance of our model on the M4 dataset. sMAPE and MAPE measure percentage-based forecast accuracy, with sMAPE normalizing by the sum of actual and forecast values. OWA combines sMAPE and MASE, benchmarking results against the naïve seasonal model, while MASE assesses error scaled by the in-sample mean absolute error. These metrics provide a comprehensive evaluation across diverse time series in the M4 dataset. Figure \ref{fig-ETTh2-compare} and \ref{fig-ECL-compare} show sample long-term predictions for best four architectures: LMS-AutoTSF, iTransformer, TimeMixer, and PatchTST. In addition to the results presented in Table \ref{table-results}, further comparison of forecasting methods, including models not listed in the table, is provided in Figure \ref{fig-comparison}. It represents a broader analysis of the methods' performance across various datasets, highlighting the advantages of our proposed LMS-AutoTSF model. Figure \ref{fig-PEMS03-compare} shows the sample prediction on PEMS03 dataset, and it is seen that proposed LMS-AutoTSF has closer predictions to the ground truth. More details for the PEMS prediction comparison are represented in the appendix.
For short-term forecasting performance, proposed method has competitive performance with TimeMixer, and outperforms iTransformer and PatchTST as shown in Table \ref{table-resultspems}. Table \ref{table-executionpems} presents a comparison of multivariate short-term forecasting models in terms of execution time (in seconds) and the number of FLOPs (floating-point operations) across four PEMS datasets. LMS-AutoTSF consistently demonstrates the fastest execution time, outperforming models like TimeMixer, iTransformer, and PatchTST. Additionally, LMS-AutoTSF also achieves significantly fewer FLOPs, highlighting its computational efficiency. The number of FLOPs was calculated using the \textit{torchprofile} library, emphasizing the model's optimized performance with reduced computational overhead compared to other methods.

\begin{table*}[hbt!]
\centering
\scriptsize
\captionsetup{justification=centering}
\caption{Comparison of multivariate long-term forecasting results with different prediction horizons (96,192,336,720) and fixed look-back 96. The values highlighted in red and underlined indicate the best MSE, while values that are only highlighted in red represent the second-best MSE. Similar for MAE highlighted in blue color.}
\tiny
\begin{tabularx}{\textwidth}{|c|c|XX|XX|XX|XX|XX|XX|X|X|X|X|}
\hline
Models       &        & \multicolumn{2}{c}{LMS-AutoTSF}                                                & \multicolumn{2}{c}{\begin{tabular}[c]{@{}c@{}}Crossformer\\\end{tabular}} & \multicolumn{2}{c}{\begin{tabular}[c]{@{}c@{}}ETSFormer\\\end{tabular}} & \multicolumn{2}{c}{\begin{tabular}[c]{@{}c@{}}iTransformer\\ \end{tabular}} & \multicolumn{2}{c}{\begin{tabular}[c]{@{}c@{}}PatchTST\\\end{tabular}} & \multicolumn{2}{c}{\begin{tabular}[c]{@{}c@{}}DLinear\\\end{tabular}} & \multicolumn{2}{c}{\begin{tabular}[c]{@{}c@{}}FEDformer\\ \end{tabular}} & \multicolumn{2}{c}{\begin{tabular}[c]{@{}c@{}}TimeMixer\\\end{tabular}} \\ \hline
Database     & Metric & MSE                                & MAE                                & MSE                                   & MAE                                                          & MSE                                              & MAE                                             & MSE                                               & MAE                                              & MSE                                                        & MAE                                & MSE                                             & MAE                                            & MSE                                                            & MAE                               & MSE                                           & MAE                                                \\ \hline
ETTh1        & 96     & 0.383                              & 0.397                              & 0.407                                 & 0.437                                                        & 0.521                                            & 0.509                                           & 0.386                                             & 0.404                                            & 0.386                                                      & 0.405                              & 0.396                                           & 0.411                                          & 0.377                                                          & 0.418                             & 0.378                                         & 0.399                                              \\
             & 192    & 0.435                              & 0.426                              & 0.461                                 & 0.462                                                        & 0.621                                            & 0.593                                           & 0.443                                             & 0.434                                            & 0.447                                                      & 0.440                              & 0.445                                           & 0.440                                          & 0.419                                                          & 0.444                             & 0.440                                         & 0.431                                              \\
             & 336    & 0.469                              & 0.442                              & 0.623                                 & 0.587                                                        & 0.721                                            & 0.692                                           & 0.492                                             & 0.463                                            & 0.489                                                      & 0.473                              & 0.487                                           & 0.465                                          & 0.459                                                          & 0.467                             & 0.499                                         & 0.459                                              \\
             & 720    & 0.478                              & 0.464                              & 0.740                                 & 0.662                                                        & 0.579                                            & 0.534                                           & 0.508                                             & 0.491                                            & 0.506                                                      & 0.496                              & 0.513                                           & 0.510                                          & 0.502                                                          & 0.503                             & 0.549                                         & 0.510                                              \\ \hline
\textbf{Avg} &        & \textcolor{red}{0.441}       & \textcolor{blue}{\underline{0.432}} & 0.557                                 & 0.537                                                        & 0.610                                            & 0.582                                           & 0.457                                             & \textcolor{blue}{0.448}                     & 0.457                                                      & 0.453                              & 0.460                                           & 0.456                                          & \textcolor{red}{\underline{0.439}}                             & 0.458                             & 0.466                                         & 0.449                                              \\ \hline
ETTh2        & 96     & 0.293                              & 0.344                              & 1.391                                 & 0.869                                                        & 0.439                                            & 0.452                                           & 0.294                                             & 0.346                                            & 0.318                                                      & 0.361                              & 0.341                                           & 0.395                                          & 0.351                                                          & 0.392                             & 0.292                                         & 0.343                                              \\
             & 192    & 0.368                              & 0.393                              & 2.616                                 & 1.258                                                        & 0.369                                            & 0.436                                           & 0.379                                             & 0.398                                            & 0.389                                                      & 0.407                              & 0.482                                           & 0.479                                          & 0.442                                                          & 0.449                             & 0.383                                         & 0.403                                              \\
             & 336    & 0.411                              & 0.427                              & 4.066                                 & 1.697                                                        & 0.498                                            & 0.451                                           & 0.434                                             & 0.435                                            & 0.428                                                      & 0.439                              & 0.593                                           & 0.542                                          & 0.498                                                          & 0.491                             & 0.439                                         & 0.437                                              \\
             & 720    & 0.432                              & 0.447                              & 3.002                                 & 1.470                                                        & 0.459                                            & 0.481                                           & 0.467                                             & 0.466                                            & 0.439                                                      & 0.454                              & 0.840                                           & 0.661                                          & 0.480                                                          & 0.487                             & 0.441                                         & 0.450                                              \\ \hline
\textbf{Avg} &        & \textcolor{red}{\underline{0.376}} & \textcolor{blue}{\underline{0.403}} & 2.768                                 & 1.324                                                        & 0.441                                            & 0.455                                           & 0.393                                             & 0.411                                            & 0.393                                                      & 0.415                              & 0.564                                           & 0.519                                          & 0.442                                                          & 0.454                             & \textcolor{red}{0.388}                  & \textcolor{blue}{0.408}                       \\ \hline
ETTm1        & 96     & 0.318                              & 0.357                              & 0.455                                 & 0.476                                                        & 0.201                                            & 0.269                                           & 0.341                                             & 0.376                                            & 0.324                                                      & 0.364                              & 0.346                                           & 0.374                                          & 0.366                                                          & 0.412                             & 0.323                                         & 0.362                                              \\
             & 192    & 0.357                              & 0.378                              & 0.487                                 & 0.502                                                        & 0.293                                            & 0.342                                           & 0.381                                             & 0.395                                            & 0.345                                                      & 0.386                              & 0.382                                           & 0.391                                          & 0.434                                                          & 0.451                             & 0.368                                         & 0.386                                              \\
             & 336    & 0.385                              & 0.403                              & 0.643                                 & 0.614                                                        & 0.337                                            & 0.391                                           & 0.419                                             & 0.418                                            & 0.395                                                      & 0.408                              & 0.415                                           & 0.415                                          & 0.484                                                          & 0.476                             & 0.401                                         & 0.406                                              \\
             & 720    & 0.450                              & 0.441                              & 0.782                                 & 0.679                                                        & 0.388                                            & 0.435                                           & 0.486                                             & 0.455                                            & 0.397                                                      & 0.409                              & 0.472                                           & 0.451                                          & 0.512                                                          & 0.489                             & 0.431                                         & 0.441                                              \\ \hline
\textbf{Avg} &        & 0.377                              & 0.394       & 0.591                                 & 0.567                                                        & \textcolor{red}{\underline{0.304}}               & \textcolor{blue}{\underline{0.359}}              & 0.406                                             & 0.411                                            & \textcolor{red}{0.365}                               & \textcolor{blue}{0.391}                              & 0.403                                           & 0.407                                          & 0.449                                                          & 0.457                             & 0.380                                         & 0.398                       \\ \hline
ETTm2        & 96     & 0.171                              & 0.254                              & 0.282                                 & 0.366                                                        & 0.189                                            & 0.280                                           & 0.184                                             & 0.266                                            & 0.186                                                      & 0.268                              & 0.193                                           & 0.293                                          & 0.202                                                          & 0.286                             & 0.175                                         & 0.256                                              \\
             & 192    & 0.236                              & 0.298                              & 0.388                                 & 0.463                                                        & 0.253                                            & 0.319                                           & 0.252                                             & 0.311                                            & 0.253                                                      & 0.311                              & 0.284                                           & 0.361                                          & 0.271                                                          & 0.329                             & 0.238                                         & 0.299                                              \\
             & 336    & 0.295                              & 0.337                              & 1.260                                 & 0.802                                                        & 0.314                                            & 0.357                                           & 0.313                                             & 0.349                                            & 0.324                                                      & 0.358                              & 0.385                                           & 0.429                                          & 0.331                                                          & 0.367                             & 0.299                                         & 0.339                                              \\
             & 720    & 0.406                              & 0.402                              & 3.257                                 & 1.246                                                        & 0.414                                            & 0.413                                           & 0.409                                             & 0.405                                            & 0.406                                                      & 0.402                              & 0.556                                           & 0.523                                          & 0.426                                                          & 0.423                             & 0.401                                         & 0.405                                              \\ \hline
\textbf{Avg} &        & \textcolor{red}{\underline{0.277}} & \textcolor{blue}{\underline{0.323}} & 1.296                                 & 0.719                                                        & 0.292                                            & 0.342                                           & 0.289                                             & 0.332                                            & 0.292                                                      & 0.334                              & 0.354                                           & 0.401                                          & 0.307                                                          & 0.351                             & \textcolor{red}{0.278}                  & \textcolor{blue}{0.324}                       \\ \hline
weather      & 96     & 0.154                              & 0.202                              & 0.174                                 & 0.239                                                        & 0.221                                            & 0.286                                           & 0.179                                             & 0.219                                            & 0.175                                                      & 0.218                              & 0.196                                           & 0.256                                          & 0.222                                                          & 0.303                             & 0.162                                         & 0.208                                              \\
             & 192    & 0.202                              & 0.247                              & 0.226                                 & 0.301                                                        & 0.269                                            & 0.302                                           & 0.225                                             & 0.258                                            & 0.222                                                      & 0.257                              & 0.239                                           & 0.299                                          & 0.289                                                          & 0.361                             & 0.207                                         & 0.251                                              \\
             & 336    & 0.259                              & 0.288                              & 0.289                                 & 0.358                                                        & 0.271                                            & 0.334                                           & 0.279                                             & 0.298                                            & 0.279                                                      & 0.298                              & 0.281                                           & 0.331                                          & 0.339                                                          & 0.382                             & 0.263                                         & 0.292                                              \\
             & 720    & 0.338                              & 0.340                              & 0.372                                 & 0.412                                                        & 0.294                                            & 0.357                                           & 0.328                                             & 0.351                                            & 0.353                                                      & 0.346                              & 0.345                                           & 0.382                                          & 0.399                                                          & 0.410                             & 0.343                                         & 0.344                                              \\ \hline
\textbf{Avg} &        & \textcolor{red}{\underline{0.238}} &  \textcolor{blue}{0.269}       & 0.265                                 & 0.327                                                        & 0.263                                            & 0.319                                           & \textcolor{blue}{\underline{0.252}}                & 0.281                                            & 0.257                                                      & 0.279                              & 0.265                                           & 0.317                                          & 0.312                                                          & 0.364                             & \textcolor{red}{0.243}                  & 0.273                                              \\ \hline
Electricity  & 96     & 0.154                              & 0.252                              & 0.154                                 & 0.255                                                        & 0.187                                            & 0.304                                           & 0.149                                             & 0.240                                            & 0.180                                                      & 0.273                              & 0.199                                           & 0.287                                          & 0.233                                                          & 0.341                             & 0.158                                         & 0.249                                              \\
             & 192    & 0.165                              & 0.260                              & 0.231                                 & 0.310                                                        & 0.199                                            & 0.315                                           & 0.164                                             & 0.255                                            & 0.194                                                      & 0.289                              & 0.201                                           & 0.290                                          & 0.271                                                          & 0.380                             & 0.169                                         & 0.259                                              \\
             & 336    & 0.177                              & 0.274                              & 0.324                                 & 0.371                                                        & 0.212                                            & 0.329                                           & 0.183                                             & 0.275                                            & 0.217                                                      & 0.322                              & 0.212                                           & 0.305                                          & 0.305                                                          & 0.386                             & 0.189                                         & 0.281                                              \\
             & 720    & 0.204                              & 0.299                              & 0.404                                 & 0.425                                                        & 0.233                                            & 0.345                                           & 0.228                                             & 0.312                                            & 0.258                                                      & 0.352                              & 0.248                                           & 0.338                                          & 0.372                                                          & 0.434                             & 0.228                                         & 0.314                                              \\ \hline
\textbf{Avg} &        & \textcolor{red}{\underline{0.175}} &  \textcolor{blue}{ 0.271}       & 0.278                                 & 0.340                                                        & \textcolor{blue}{\underline{0.207}}               & 0.323                                           & \textcolor{red}{0.181}                      & \textcolor{blue}{0.270}                     & 0.212                                                      & 0.309                              & 0.215                                           & 0.305                                          & 0.295                                                          & 0.385                             & 0.186                                         & 0.275                                              \\ \hline
Traffic      & 96     & 0.485                              & 0.330                              & 0.520                                 & 0.277                                                        & 0.607                                            & 0.392                                           & 0.417                                             & 0.287                                            & 0.526                                                      & 0.347                              & 0.696                                           & 0.428                                          & 0.590                                                          & 0.372                             & 0.469                                         & 0.291                                              \\
             & 192    & 0.483                              & 0.322                              & 0.569                                 & 0.304                                                        & 0.621                                            & 0.399                                           & 0.432                                             & 0.292                                            & 0.522                                                      & 0.332                              & 0.650                                           & 0.409                                          & 0.608                                                          & 0.383                             & 0.480                                         & 0.299                                              \\
             & 336    & 0.493                              & 0.323                              & 0.575                                 & 0.315                                                        & 0.622                                            & 0.396                                           & 0.449                                             & 0.305                                            & 0.531                                                      & 0.339                              & 0.657                                           & 0.412                                          & 0.618                                                          & 0.385                             & 0.498                                         & 0.301                                              \\
             & 720    & 0.524                              & 0.336                              & 0.589                                 & 0.321                                                        & 0.632                                            & 0.396                                           & 0.481                                             & 0.318                                            & 0.552                                                      & 0.352                              & 0.679                                           & 0.428                                          & 0.646                                                          & 0.395                             & 0.506                                         & 0.313                                              \\ \hline
\textbf{Avg} &        & 0.497                              & 0.328                              & 0.563                                 & \textcolor{blue}{0.304}                                 & 0.620                                            & 0.395                                           & \textcolor{red}{\underline{0.444}}                & \textcolor{blue}{\underline{0.301}}               & 0.532                                                      & 0.342                              & 0.670                                           & 0.419                                          & 0.615                                                          & 0.383                             & \textcolor{red}{0.488}                  & \textcolor{blue}{\underline{0.301}}                 \\ \hline
Exchange     & 96     & 0.082                              & 0.199                              & 0.256                                 & 0.367                                                        & 0.085                                            & 0.204                                           & 0.098                                             & 0.226                                            & 0.097                                                      & 0.216                              & 0.098                                           & 0.232                                          & 0.166                                                          & 0.294                             & 0.088                                         & 0.208                                              \\
             & 192    & 0.174                              & 0.296                              & 0.469                                 & 0.508                                                        & 0.182                                            & 0.303                                           & 0.188                                             & 0.312                                            & 0.182                                                      & 0.303                              & 0.186                                           & 0.325                                          & 0.281                                                          & 0.387                             & 0.184                                         & 0.307                                              \\
             & 336    & 0.324                              & 0.410                              & 0.896                                 & 0.740                                                        & 0.348                                            & 0.428                                           & 0.337                                             & 0.422                                            & 0.346                                                      & 0.425                              & 0.327                                           & 0.434                                          & 0.469                                                          & 0.502                             & 0.376                                         & 0.442                                              \\
             & 720    & 0.834                              & 0.686                              & 1.401                                 & 0.966                                                        & 1.025                                            & 0.774                                           & 0.885                                             & 0.714                                            & 0.950                                                      & 0.731                              & 0.749                                           & 0.663                                          & 1.164                                                          & 0.828                             & 0.880                                         & 0.701                                              \\ \hline
\textbf{Avg} &        & \textcolor{red}{\underline{0.353}} & \textcolor{blue}{\underline{0.398}} & 0.755                                 & 0.645                                                        & 0.410                                            & 0.427                                           & \textcolor{red}{0.377}                      & 0.418                                            & 0.393                                                      & 0.418                              & 0.340                                           & 0.413                                          & 0.520                                                          & 0.502                             & 0.382                                         & \textcolor{blue}{0.414}                       \\ \hline
\end{tabularx}
\label{table-results}
\end{table*}

\begin{table*}[hbt!]
\centering
\scriptsize
\captionsetup{justification=centering}
\caption{Short-term forecasting results in the M4 dataset with a single variate. All prediction lengths are in [6, 48].  The values highlighted in red indicate the best results for each row.}
\adjustbox{max width=\textwidth}{
\begin{tabular}{|l|c|c|c|c|c|}
\textbf{Metric}                 & \textbf{Category} & \textbf{LMSAutoTSF}                                                  & \textbf{iTransformer} & \textbf{TimeMixer}                                                   & \textbf{PatchTST} \\ \hline
\multirow{5}{*}{\textbf{sMAPE}} & Yearly            & \textcolor{red}{13.410}    & 14.409  & 13.358 & 13.588     \\
                                & Quarterly         & \textcolor{red}{10.056}  & 10.777  & 10.143  & 10.909      \\
                                & Monthly           & \textcolor{red}{12.763}  & 16.650 & 12.797  & 14.149    \\
                                & Others            & \textcolor{red}{4.940}   & 5.543   & 5.046  & 5.853 \\ \hline
                                & \textbf{Average}  & \textcolor{red}{\textbf{11.871}} & \textbf{14.170}  & \textbf{11.902}                              & \textbf{12.827}   \\ \hline
\multirow{5}{*}{\textbf{MAPE}}  & Yearly            & 16.279   & 19.191 & \textcolor{red}{16.257} & 16.709    \\
                                & Quarterly         & \textcolor{red}{11.552}   & 12.871  & 11.591 & 12.924 \\
                                & Monthly           & 14.988  & 20.144 & \textcolor{red}{14.961} & 16.873  \\
                                & Others            & 6.667 & 7.750 & \textcolor{red}{6.573} & 10.478 \\ \hline
                                & \textbf{Average}  & \textbf{14.044}                                                      & \textbf{17.560}       & \textcolor{red}{\textbf{14.031}} & \textbf{15.568}   \\ \hline 
\multirow{5}{*}{\textbf{MASE}}  & Yearly            & 3.031   & 3.218 & \textcolor{red}{3.028}   & 3.036   \\
                                & Quarterly         & \textcolor{red}{1.178}   & 1.284   & 1.193  & 1.315    \\
                                & Monthly           & \textcolor{red}{0.934}   & 1.392  & 0.942  & 1.122             \\
                                & Others            & \textcolor{red}{3.262}    & 3.998   & 3.398 & 3.782             \\ \hline
                                & \textbf{Average}  & \textcolor{red}{\textbf{1.591}}   & \textbf{1.916} & \textbf{1.605}  & \textbf{1.742}    \\ \hline
\multirow{5}{*}{\textbf{OWA}}   & Yearly            & 0.792  & 0.846  & \textcolor{red}{0.790} & 0.798    \\
                                & Quarterly         & \textcolor{red}{0.886}  & 0.957   & 0.896 & 0.975   \\
                                & Monthly           & \textcolor{red}{0.881}  & 1.232  & 0.887  & 1.018    \\
                                & Others            & \textcolor{red}{1.034}   & 1.214  & 1.067 & 1.212   \\ \hline
                                & \textbf{Average}  & \textcolor{red}{\textbf{0.854}}  & \textbf{1.023}  & \textbf{0.858}  & \textbf{0.928}    \\ \hline
\end{tabular}
}
\label{table-resultsM4}
\end{table*}

%\vskip 0pt plus -1fil

\begin{table*}[hbt!]
\centering
\scriptsize
\caption{Comparison of multivariate short-term forecasting results with different prediction horizons (12,24,48) and fixed look-back 96. The values highlighted in red and underlined indicate the best MSE, while values that are only highlighted in red represent the second best MSE. Similar for MAE highlighted in blue colour}
\begin{tabular}{|c|c|cc|cc|cc|cc|}
\hline
{Models}                            & {\color[HTML]{C9211E} } & \multicolumn{2}{c}{{LMS-AutoTSF}} & \multicolumn{2}{c}{iTransformer} & \multicolumn{2}{c}{TimeMixer} & \multicolumn{2}{c}{PatchTST}                                  \\
\hline
\textit{Dataset}                  & Metric                  & MSE                        & MAE                       & MSE             & MAE            & MSE           & MAE           & MSE                           & MAE                           \\
\hline
                                 & 12                      & 0.0647                     & 0.1706                    & 0.0732          & 0.1804         & 0.0607        & 0.1659        & 0.0915                        & 0.2018                        \\
                                & 24                      & 0.0776                     & 0.1898                    & 0.1013          & 0.2139         & 0.0822        & 0.1939        & 0.1457                        & 0.2573                        \\
                               & 48                      & 0.1022                     & 0.2191                    & 0.1663          & 0.2789         & 0.0982        & 0.2138        & 0.3014                        & 0.3791                        \\
\hline
\multirow{-4}{*}{{ \textit{PEMS03}}} & Avg                     & \textcolor{red}{0.0815}                     & \textcolor{blue}{0.193}                     & 0.1136          & 0.2244         & \textcolor{red}{\underline{0.0803}}        & \textcolor{blue}{\underline{0.1912}}        & 0.179                         & 0.2794                        \\
\hline
{ }                                  & 12                      & 0.0664                     & 0.1681                    & 0.0868          & 0.1925         & 0.0664        & 0.1689        & 0.1075                        & 0.2262                        \\
{ }                                  & 24                      & 0.0749                     & 0.1813                    & 0.1221          & 0.2313         & 0.0746        & 0.1803        & 0.1725                        & 0.2817                        \\
{ }                                  & 48                      & 0.0862                     & 0.1975                    & 0.1935          & 0.2977         & 0.0905        & 0.2039        & 0.3845                        & 0.4423                        \\
\hline
\multirow{-4}{*}{{\textit{PEMS04}}} & Avg                     & \textcolor{red}{\underline{0.0758}}                     & \textcolor{blue}{\underline{0.1823}}                    & 0.134           & 0.2405         & \textcolor{red}{0.0771}        & \textcolor{blue}{0.184}         & 0.2215                        & 0.316                         \\
\hline
{ }                                  & 12                      & 0.0565                     & 0.1582                    & 0.0711          & 0.1739         & 0.0555        & 0.1538        & { 0.098}  & { 0.217}  \\
{ }                                  & 24                      & 0.0715                     & 0.1815                    & 0.1039          & 0.21           & 0.0661        & 0.1688        & 0.1486                        & 0.2628                        \\
{ }                                  & 48                      & 0.0928                     & 0.2091                    & 0.1656          & 0.2695         & 0.0886        & 0.2013        & 0.3215                        & 0.3941                        \\
\hline
\multirow{-4}{*}{{ \textit{PEMS07}}} & Avg                     & \textcolor{red}{0.0736}                     & \textcolor{blue}{0.1829}                    & 0.113           & 0.2178         & \textcolor{red}{\underline{0.070}}         & \textcolor{blue}{\underline{0.174}}         & 0.189                         & 0.2913                        \\
\hline
{ }                                  & 12                      & 0.064                      & 0.1699                    & 0.0817          & 0.1881         & 0.0617        & 0.1647        & { 0.1492} & { 0.2125} \\
{ }                                  & 24                      & 0.0798                     & 0.1948                    & 0.122           & 0.2347         & 0.0779        & 0.1872        & 0.1639                        & 0.2798                        \\
                               & 48                      & 0.1039                     & 0.227                     & 0.2217          & 0.3279         & 0.1015        & 0.2197        & 0.3126                        & 0.3934                        \\
\hline
\multirow{-4}{*}{{ \textit{PEMS08}}} & Avg                     & \textcolor{red}{0.082}                      & \textcolor{blue}{0.197}                     & 0.1418          & 0.250          & \textcolor{red}{\underline{0.080}}         & \textcolor{blue}{\underline{0.190}}         & 0.208                         & 0.295  \\
\hline
\end{tabular}
\label{table-resultspems}
\end{table*}

%\vskip 0pt plus -1fil

\begin{table*}[hbt!]
\centering
\scriptsize
\caption{Comparison of multivariate short-term forecasting results in terms of execution time (sec) and no. of flops}
\begin{tabular}{|c|l|c|c|c|c|}
\hline
\multicolumn{1}{l}{}           & Datasets & LMS-AutoTSF & TimeMixer & iTransformer & PatchTST \\
\hline
\multirow{4}{*}{\begin{tabular}[c]{@{}c@{}}Execution Time (Sec) \\ (For test data)\end{tabular}} & PEMS03   & \textcolor{red}{0.8233}      & 0.893     & 1.0696       & 1.3208   \\
                                      & PEMS04   & \textcolor{red}{0.7945}      & 0.8608    & 1.1218       & 1.6141   \\
                                      & PEMS07   & \textcolor{red}{0.9783}      & 1.0231    & 1.4294       & 2.2553   \\
                                      & PEMS08   & \textcolor{red}{0.7762}      & 0.8142    & 0.8227       & 0.9495   \\
                                      \hline
\multirow{4}{*}{No. of Flops}         & PEMS03     & \textcolor{red}{151.52M}      & 278.96M    & 4581.79M      & 13809.65M \\
                                      & PEMS04     & \textcolor{red}{129.93M}      & 270.17M    & 15604.76M     & 46861.69M \\
                                      & PEMS07     & 373.72M      & \textcolor{red}{369.48M}    & 5721.62M      & 34061.23M \\
                                      & PEMS08     & \textcolor{red}{71.95M}       & 246.55M    & 1102.88M      & 6557.65M \\
                                      \hline
\end{tabular}
\label{table-executionpems}
\end{table*}

%\vskip 0pt plus -1fil

\section{Conclusion}
\label{sec-conclusion}
In this work, we have introduced LMS-AutoTSF, a novel time series forecasting architecture that combines learnable decomposition with integrated autocorrelation and multi-scale processing. By leveraging filtering through learnable parameters, the model dynamically isolates trend and seasonal components directly in the frequency domain. Integrated autocorrelation further enhances the model’s ability to capture temporal dependencies, improving forecasting accuracy across various time horizons. LMS-AutoTSF consistently achieves state-of-the-art performance across a wide range of benchmarks, showcasing its generality and robustness for time series forecasting tasks.

%\clearpage

%\subsubsection*{Author Contributions}
%Conceptualization: Ibrahim Delibasoglu, Sanjay Chakraborty; Methodology: Ibrahim Delibasoglu; Formal analysis and investigation: Ibrahim Delibasoglu, Sanjay Chakraborty; Data curation: Ibrahim Delibasoglu, Sanjay Chakraborty; Software: Sanjay Chakraborty; Visualization: Ibrahim Delibasoglu, Sanjay Chakraborty; Validation: Ibrahim Delibasoglu, Sanjay Chakraborty; Writing - original draft preparation: Ibrahim Delibasoglu, Sanjay Chakraborty; Writing - review and editing: Ibrahim Delibasoglu, Sanjay Chakraborty, Fredrik Heintz; Resources: Ibrahim Delibasoglu, Sanjay Chakraborty; Project administration: Fredrik Heintz; Supervision: Fredrik Heintz. All authors have read and approved the final manuscript and have agreed to its publication.  

%\subsubsection*{Acknowledgments}
%The authors declare that they have no known competing
%financial interests or personal relationships that could have
%appeared to influence the work reported in this paper. There
%is no new data created or analysed in this study. There is no
%funding received.

\clearpage

\bibliographystyle{unsrt}
\bibliography{main}

\vfill
%\appendix
\end{document}